\documentclass[conference]{IEEEtran}
\IEEEoverridecommandlockouts
\usepackage{cite} 
\usepackage{amsmath,amssymb,amsfonts}

\usepackage{amsthm}

\DeclareMathOperator{\EE}{\mathbb{E}}
\DeclareMathOperator{\RR}{\mathbb{R}}
\DeclareMathOperator*{\argmax}{argmax}
\DeclareMathOperator*{\dom}{dom}
\DeclareMathOperator{\KS}{KS}

\theoremstyle{definition}
\newtheorem{definition}{Definition}[section]
\newtheorem{fact}[definition]{Fact}

\usepackage{tikz}
\usetikzlibrary{positioning}
\usepackage{pgfplots}

\usepackage[inline]{enumitem}

\usepackage{booktabs} 

\usepackage{algorithmic}
\usepackage{graphicx}
\usepackage{textcomp}
\usepackage{xcolor}
\def\BibTeX{{\rm B\kern-.05em{\sc i\kern-.025em b}\kern-.08em
    T\kern-.1667em\lower.7ex\hbox{E}\kern-.125emX}}
\begin{document}

\title{Machine Unlearning: Linear Filtration for Logit-based Classifiers\\
}


\author{\IEEEauthorblockN{Thomas Baumhauer}
\IEEEauthorblockA{
\textit{St. P\"olten University of Applied Sciences}\\
St. P\"olten, Austria \\
thomas.baumhauer@fhstp.ac.at}
\and
\IEEEauthorblockN{Pascal Sch\"ottle}
\IEEEauthorblockA{
\textit{Management Center Innsbruck}\\
Innsbruck, Austria \\
pascal.schoettle@mci.edu}
\and
\IEEEauthorblockN{Matthias Zeppelzauer}
\IEEEauthorblockA{
\textit{St. P\"olten University of Applied Sciences}\\
St. P\"olten, Austria \\
matthias.zeppelzauer@fhstp.ac.at}
}

\maketitle

	\begin{abstract}		
	Recently enacted legislation grants individuals certain rights to decide
	in what fashion their personal data may be used and in particular a
	``right to be forgotten''. This poses a challenge to machine learning:
	how to proceed when an individual retracts permission to use data
	which has been part of the training process of a model?
	From this question emerges the field of {\em machine unlearning},
	which could be broadly described as the investigation of 
	how to ``delete training data from models''.
	Our work complements this direction of research for the specific setting of class-wide deletion requests for
	classification models (e.g. deep neural networks).
	As a first step, we propose {\em linear filtration} as an intuitive, computationally efficient sanitization method. Our experiments demonstrate benefits in an
	adversarial setting over naive deletion
	schemes.
\end{abstract}


	\section{Introduction}
Recently enacted legislation, such as the
European Union’s General Data Protection Regulation (GDPR) \cite{GDPR},
previously its ``right to be forgotten'' \cite{RTBFG},
and the California Consumer Privacy Act~\cite{CCPA} grant individuals certain rights to decide in what fashion their personal
data may be used, and in particular the right to ask for personal data collected about them to be deleted.

At present the implementation of such rights in the context of machine learning models trained on personal
data is largely an open problem.
In \cite{Villaronga2017} the authors even conclude that
``it may be impossible to fulfill the legal aims of the Right to be Forgotten in artificial intelligence environments''.

Indeed, machine learning models may unintentionally memorize (part of) their training data,
leading to privacy issues in many applications, e.g. image classification \cite{Yeom2018,Sablayrolles2019}
or natural language processing \cite{Carlini2019}, and potentially enabling an adversary to extract
sensitive information from a trained model by so-called {\em model inversion} \cite{Veale2018}.

\begin{figure}				\begin{center}
		\centerline{\includegraphics[width=\columnwidth]{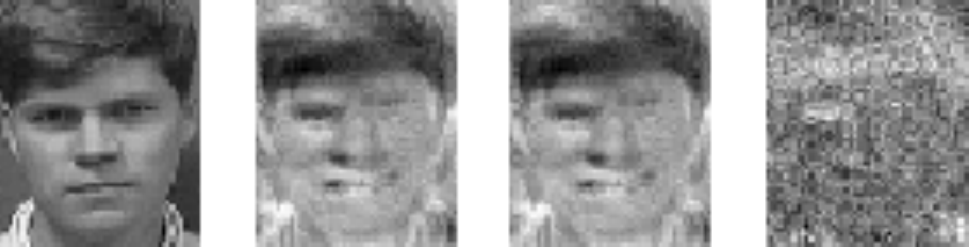}}
	\end{center}
	\vskip -0.2in
	\caption{
		Results of a model inversion attack for a toy model trained on
		the AT\&T Faces dataset with 4 classes.
		For one of the classes, from left to right:
		one of the training images, reconstruction
		of the class by model inversion, reconstruction
		after naive unlearning, reconstruction after
		unlearning by our proposed method of
		{\em normalizing linear filtration} (defined in section~\ref{sec:method}).
		The reconstructions of the other classes remain visually unchanged
		by normalizing linear filtration, see figure \ref{fig:att_inversion2}.
	}
	\label{fig:att_inversion1}
	\vskip 0.2in
\end{figure}

Informally, deletion of part of the training data from a machine learning model can be understood as
removal of its influence on the model parameters, in order to obtain
a model that ``looks as if it has never seen that part of the data''.
We refer to this process as {\em unlearning}.	
Clearly, the problem of unlearning can be solved in a trivial fashion, by simply retraining the model without using
the	part of the data we wish to unlearn. For large, real-world models, retraining from scratch 
may incur significant computational costs, and may thus be practically infeasible, if deletion requests
are expected to arrive at any time.	
We therefore wish to find more efficient unlearning algorithms,
which is notoriously difficult, owing to the fact that
for many popular learning algorithms every training data point potentially affects every model parameter.

First approaches towards directed unlearning were introduced in \cite{Cao2015},
and more recently in \cite{Ginart2019,bourtoule2019}.
In our work we consider the problem of unlearning in the setting of classification models for which,
in contrast to previous work, we assume that single individuals own all training data associated with
a particular class, as may be the case e.g. in biometric applications.

In this setting, we consider classifiers that predict logits, i.e. (rescaled) logarithmic probabilities
that a data point belongs to certain classes. For such classifiers we propose a novel sanitization method that
applies a linear transformation to these predictions. For an appropriate hypothesis class
this transformation can be absorbed into the original classifier.  The computation of the transformation requires
barely more than computing predictions by the original classifier for a (small) number of data points per class.
We call this method {\em linear filtration}.
Figure \ref{fig:att_inversion1} shows
the results of this method when used as a defense against model inversion \cite{Fredrikson2015}.

In summary, the main contributions of our work are:
\begin{enumerate}[label=\textbf{\arabic*}), topsep=0pt,itemsep=6pt,partopsep=0pt, parsep=0pt]
	\item
	We develop {\em linear filtration}, a novel algorithm for the sanitization of
	classification models that predict logits, after class-wide deletion requests.
	\item 
	On the theoretical side, we add to the definition
	of unlearning in the sense of \cite{Ginart2019}, by proposing a weakened, ``black-box'' variant of the definition, which may serve as a more realistic goal
	in practice.
	\item
	As practical methodology, we suggest that the quality of an empirical unlearning operation may be evaluated in an adversarial setting, i.e.
	by testing how well it prevents certain privacy attacks on machine learning models.
\end{enumerate}

The rest of this paper is organized as follows:
Section \ref{sec:related_work} gives an overview of work related to machine unlearning, as well the
adversarial methodology employed in our experiments.
Section \ref{sec:problem_def} formalizes the unlearning of training data from machine learning models.
Section \ref{sec:method} establishes several variants of our method of linear filtration.
Section \ref{sec:experiments} experimentally evaluates linear filtration, primarily in an adversarial context.
Section \ref{sec:on_the_def} features discussion on the definition of unlearning. 

\section{Related Work}
\label{sec:related_work}

{\bf Machine unlearning. $~$}
The term {\em machine unlearning} first appears in \cite{Cao2015}. There the authors
consider unlearning the framework of statistical query learning \cite{Kearns1998}. 
This allows them to unlearn data points for learning algorithms where all queries to the data
are decided upfront. However, many popular learning algorithms (such as gradient descent)
query data adaptively. In the adaptive setting the approach of \cite{Cao2015} does not
give any bounds and quickly falls apart if the number of queries is large, as is the case
for neural networks.

\cite{Ginart2019} features a discussion of the problem of efficient unlearning  of training data points from models, establishes several engineering principles, and on the practical side
proposes unlearning algorithms for $k$-means clustering.
In particular, they recognize that given the stochastic nature of many learning algorithms
a probabilistic definition of unlearning (there ``deletion'') is necessary. We adopt this view in our work.

\cite{bourtoule2019} propose a framework they refer to as SISA (sharded, isolated, sliced, and aggregated training),
which can be thought of as bookkeeping method seeking to limit and keep track of the influence of training data points on model parameters,
thus reducing the amount of retraining necessary upon receiving a deletion request.
This approach comes at the cost of a large storage overhead.

\cite{guo2019} define {\em $\epsilon$-certified removal},
``a very
strong theoretical guarantee that a model from
which data is removed cannot be distinguished
from a model that never observed the data to begin
with'', a concept akin to that of differential privacy \cite{d2006}. Combining this with the idea of influence functions \cite{koh2017}, they then develop a certified removal mechanism for linear classifiers.

\cite{golatkar2019, golatkar2020} adopt an information theoretic view of unlearning and develop unlearning operations based on linearized model dynamics (drawing inspiration from the {\em neural tangent kernel} \cite{Jacot2018, Lee2019}, a technique to describe the
gradient descent dynamics of the training of neural networks using kernel methods).

The work done in \cite{guo2019, golatkar2019, golatkar2020} may be considered complementary to the method of {\em linear filtration} we are going to develop in this paper,
in the sense that they use strong assumptions (in particular work with linear/linearized models) obtaining stronger guarantees, while linear filtration is an intuitive heuristic, largely agnostic to model architecture.

\cite{sommer2020} propose a formal framework for verification of machine unlearning, based on machine learning backdoor attacks.

{\bf Membership inference. $~$}		
It is an open problem to find a suitable measure for the quality of unlearning, when employing a heuristic unlearning operation
with no or weak theoretical guarantees, i.e. to quantify the remaining influence of ``deleted'' training data
on a model's parameters.
In our experiments we thus take ideas from {\em membership inference}. The goal of membership inference is to
determine whether a given data point has been used in the training process of a given model. A few recent works on membership
inference include
\cite{Shokri2017, Yeom2018, Hayes2019, Sablayrolles2019}.	
In particular, our adversarial setup in section \ref{sec:evaluation} draws a large amount of inspiration from
\cite{Shokri2017}, where a binary classifier is trained on the outputs of so-called {\em shadow models}
to decide membership.	

{\bf Model inversion. $~$}
Broadly, {\em model inversion} may be defined as drawing inferences about private training data
from the outputs of a model trained on this data.	
The term was introduced in \cite{f2014}.
\cite{Fredrikson2015} reconstruct human-recognizable images of individuals
from neural networks trained for face recognition,
using gradient ascend on the input space.

We remark that model inversion is at its core the result of a correlation between
input and output space that is simply captured by the model
(and may exist independently of the model), and thus does not necessarily constitute
a privacy breach. \cite{McSherry2016} features a highly recommended elaboration of this point in much detail.

\cite{Shokri2017} conclude their discussion of model inversion with the statement
that ``model inversion produces the average of the
features that at best can characterize an entire output class.''
Thus, model inversion is of some interest in the specific context of our paper, which focuses on class-wide unlearning (and
hence implicitly makes the assumption that a single individual owns all data for an entire output class).
We experiment with model inversion in section \ref{sec:model_inversion}, see also figure \ref{fig:att_inversion1}.

{\bf Differential privacy. $~$}
Differential privacy \cite{d2006,d2014,Abadi2016}  limits the influence of individual training points
on a model in a precise probabilistic way.
We briefly remark that (differential) privacy and data deletion may be considered orthogonal problems, in the sense that
private models need not support efficient deletion, and models supporting efficient deletion need not be private.
\cite{Ginart2019} discuss this point in additional detail.

\section{Problem Definition. $~$}
\label{sec:problem_def}

In this section we formalize our notion of unlearning.

{\bf Notation. $~$}
For a vector $v \in \RR^k$ we use indices ranging from $0$ to $(k-1)$
to denote its entries, i.e. $v = (v_0, v_1, \dots v_{k-1})^\top$.
One dimensional vector are always column vectors.

For a vector of logits $\ell \in \RR^k$ let
$$
\sigma_{i^*}(\ell) = \frac{\exp({\ell_{i^*}})}
{\sum_{i<k} \exp({\ell_i})} \in [0,1], \quad \text{for all } i^* < k
$$
and $\sigma(\ell) = ( \sigma_0(\ell), \sigma_1(\ell), \dots, \sigma_{k-1}(\ell))$. We call $\sigma$ the {\em softmax function}.

We use uppercase, boldface letters for random variables. If
$\mathbf R$ is a random variable $P(\mathbf R)$ denotes its distribution.

{\bf Classification. $~$}
We consider a multiclass classification problem: Let $\mathbf X$ be a random variable taking values in some input space $\mathcal X$ (e.g.
$\mathcal X = \RR^{28 \times 28}$), and let $\mathbf Y$ be
a random variable representing class labels taking values in $\mathcal Y = \{0, 1, \dots, k-1 \}$ for some natural number $k > 2$, with some joint data generating distribution $P(\mathbf X, \mathbf Y)$.

Given $x \in \mathcal X$, a classifier $h: \mathcal X \to \RR^k$ for this classification problem attempts to
estimate logits $\ell = h(x)$ such that
$$
\sigma(\ell) \approx P(\mathbf Y \mid \mathbf X = x).
$$
{\bf Hypothesis class. $~$}
In this paper we consider the class~$\mathcal H$ of all classifiers $h$ of the form
$$
h = \mathrm{logistic\ regression} \circ \mathrm{feature\ extraction},
$$ 
i.e. classifiers that can be decomposed into a (potentially non-linear) feature extraction followed by a multinomial logistic regression.

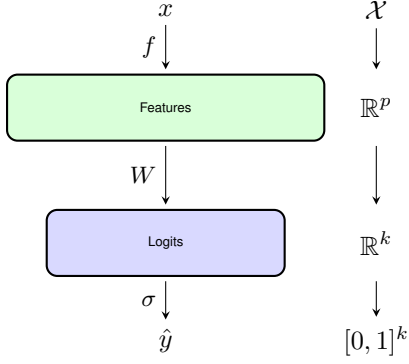
\begin{figure}
	\vskip 0.1in
	\begin{center}
		\begin{tikzpicture}

		\draw (2.8,1.3) node (in_) {$\mathcal X$};
		\draw (2.8,0) node[minimum height=2.5em, outer sep=2pt] (nn_) {$\RR^p$};
		\draw (2.8,-1.8) node[minimum height=2.5em, outer sep=2pt] (dense_) {$\RR^k$};
		\draw (2.8,-3.1) node (out_) {$[0,1]^k$};

		\draw (0,1.3) node (in) {$x$};

		\draw (0,0) node[draw,thick,fill=green!15, rounded corners,font=\sffamily\tiny, minimum height=2.5em, minimum width=12em,outer sep=2pt] (nn) {Features};
		
		\draw (0, -1.8) node[draw,thick,fill=blue!15, rounded corners, font=\sffamily\tiny,minimum height=2.5em, minimum width=9em, outer sep=2pt] (dense) {Logits};
		
		\draw (0,-3.1) node (out) {${\hat y}$};
		
		\draw[->, black!100, >=stealth] (in) -- (nn) node[midway,left,black]
		{\small $f$};
		\draw[->, black!100, >=stealth] (nn) -- (dense) node[midway,left,black]
		{\small $W$};
		\draw[->, black!100, >=stealth] (dense) -- (out) node[midway,left,black]
		{\small $\sigma$};
		
		\draw[->, black!100, >=stealth] (in_) -- (nn_);
		\draw[->, black!100, >=stealth] (nn_) -- (dense_);
		\draw[->, black!100, >=stealth] (dense_) -- (out_);
		
		\end{tikzpicture}
		\caption{Schematic representation of
			$\sigma \circ h$, for a classifier $h~=~W \circ f$ in the hypothesis class
			considered throughout this paper.
			Here $f$ denotes a feature extraction,
			$W$ is a linear transformation and $\sigma$ is the softmax function.
			In deep learning terminology ``Logits'' represents
			a fully connected layer with $k$~units and weights~$W$.
		}
		\label{fig:schematic_h}
	\end{center}
	\vskip 0.2in
\end{figure}

More formally any $h \in \mathcal H$ can be expressed as
\begin{align}
	\label{form:decomposition}
	h: x \mapsto W \cdot f(x)
\end{align}
where $f : \mathcal X \to \RR^p$ denotes the feature extraction, $p$ denotes the dimension of the feature space, $W$ is a $(k \times p)$-matrix
representing a linear transformation $\RR^p \to \RR^k$, and $\sigma$
denotes the softmax function. Figure \ref{fig:schematic_h} shows a schematic representation of the elements of $\mathcal H$. To simplify notation we do not consider affine transformations, i.e. classifiers of the form
$h: x \mapsto W \cdot f(x) + b$ for some $b \in \RR^k$. However, we remark that our method is easily adapted to this case.

Observe that in particular all deep neural networks for which the output is a densely connected layer with softmax activations fit into the schema discussed in
this section.

{\bf Learning algorithms. $~$}
For a finite training set $D~\subseteq~\mathcal X~\times~\mathcal Y$, a learning algorithm
$\mathbf A$ calculates a classifier
$$h = \mathbf A(D) \in \mathcal H.$$
Note that if $\mathbf A$ is non-deterministic $\mathbf A(D)$ can be considered a random variable
taking values in $\mathcal H$.

{\bf Unlearning of classes. $~$}
Let $\mathcal C \subsetneq \mathcal Y$ be a set of classes, 
which we want to unlearn.
Consider the multiclass classification problem for the distribution $P(\mathbf X, \mathbf Y \mid \mathbf Y \not \in \mathcal C)$, i.e. the original problem with the
classes $\mathcal C$ removed.
We define the hypothesis class $\mathcal H_{\lnot \mathcal C}$ for this problem similarly to $\mathcal H$, i.e.
$h \in \mathcal H_{\lnot \mathcal C}$ is of form (\ref{form:decomposition}) with $W$ a $((k - |\mathcal C|) \times p)$-Matrix.
For $D \subseteq \mathcal X \times \mathcal Y$ let $D_{\lnot \mathcal C} =
\{(x,y) \in D : y \not \in \mathcal C \}$ and let $\mathbf A_{\lnot \mathcal C}(D_{\lnot \mathcal C})
\in \mathcal H_{\lnot \mathcal C}$ denote a
classifier calculated by some
learning algorithm $\mathbf A_{\lnot \mathcal C}$.

\begin{definition}[Unlearning]
	\label{dfn:deletion}
	We say that a map
	$$
	\mathfrak D: \mathcal H \to \mathcal H_{\lnot \mathcal C}
	$$
	``unlearns $\mathcal C$ from $\mathcal H$ with respect to $\mathbf A,
	\mathbf A_{\lnot \mathcal C}, D$''
	if the random variables $\mathfrak D(\mathbf A(D))$ and
	$\mathbf A_{\lnot \mathcal C}(D_{\lnot \mathcal C})$ have the same distribution over $\mathcal H_{\lnot \mathcal C}$. We call $\mathfrak D$ an {\em unlearning operation} (for $\mathcal C$).
\end{definition}

{\bf Weak unlearning of classes. $~$}
A good choice of $\mathfrak D$ will of course depend on the learning algorithm $\mathbf A$.
We mostly concern ourselves with the case where $\mathbf A$ trains
a neural network with a densely connected output layer with softmax activations. Unfortunately it is difficult to
understand how a neural network represents knowledge internally (e.g. \cite{aless2019}), hence unlearning as defined above may currently be out of reach.
We therefore propose a weakening of the
above definition.

\begin{definition}[Weak unlearning]
	\label{dfn:weak_deletion}
	As before let $\mathfrak D$ be a map $\mathcal H \to \mathcal H_C$
	and for $\mathbf X$ (taking values in the input space
	$\mathcal X$ according to the data generating distribution) consider the random variables
	\begin{align*}
		\mathbf L_{\mathrm{seen}} = h_0(\mathbf X),
		&\quad \text{where } h_0 = \mathfrak D (\mathbf A(D))\\
		\mathbf L_{\lnot \mathrm{seen}}= h_1(\mathbf X),
		&\quad \text{where } h_1 = \mathbf A_{\lnot \mathcal C}(D_{\lnot \mathcal C})
	\end{align*}
	i.e. the logit outputs of the respective classifiers, 
	taking values in $\RR^{k - |\mathcal C|}$.
	We say that
	``$\mathfrak D$ weakly unlearns $\mathcal C$ from $\mathcal H$ with respect to $\mathbf A, \mathbf A_{\lnot \mathcal C}, D$''
	if $\mathbf L_{\mathrm{seen}}$ and $\mathbf L_{\lnot \mathrm{seen}}$ have the same distribution
	over $\RR^{k - |\mathcal C|}$. We call $\mathfrak D$ a {\em weak unlearning operation} (for $\mathcal C$).
\end{definition}

\begin{fact}
	If $\mathfrak D$ is an unlearning operation, then
	$\mathfrak D$ is a weak unlearning operation. \qed
\end{fact}

{\bf Discussion. $~$}
Definition \ref{dfn:deletion} demands that the distributions over
the hypothesis class (i.e. in practical terms the parameter space)
are the same, while \ref{dfn:weak_deletion} relaxes this to the distributions
over the output space being the same. Intuitively speaking: \ref{dfn:deletion} demands $\mathfrak D$ make the model $h$ ``look as if $h$ had never seen the data'', while \ref{dfn:weak_deletion} demands
$\mathfrak D$ make the {\bf outputs} of $h$ ``look as if $h$ had never seen the data''
We may therefore consider
unlearning (in the strong sense) to be a white-box variant 
and unlearning in the weak sense to be a black-box variant of the definition
of unlearning. See section \ref{sec:on_the_def} for further discussion.

Abusing the terminology introduced in this section we refer to maps $\mathfrak D$
that roughly fit definitions \ref{dfn:deletion} and \ref{dfn:weak_deletion}, respectively in the sense
that they make the relevant distributions similar in an appropriate divergence measure,
as ``good'', ``satisfactorily performing'', etc. unlearning operations.

\section{Method}
\label{sec:method}
In this section we propose an intuitive weak unlearning operation $\mathfrak D$ for classes,
exploiting the special structure of elements of $\mathcal H$.
In our experiments in section \ref{sec:experiments} we demonstrate that $\mathfrak D$ performs satisfactorily for neural networks.

\subsection{Definition of weak unlearning operation $\mathfrak D_z$}
\label{sec:weak_del_op}
For simplicity of notation let $\mathcal C = \{0\}$, i.e. we are going to unlearn class $0$. However our method easily generalizes to arbitrary $\mathcal C$. Let $h = W \circ f \in \mathcal H$ be a classifier.
For $j < k$ let
$$
a_j = \EE[
h(\mathbf X)
\mid \mathbf Y = j] \in \RR^{k}
$$
be the expected prediction for class $j$, and let
$$
A = \Big(a_0 \mid a_1 \mid \dots \mid a_{k-1}\Big) \in \RR^{k \times k}.
$$
In practice, we may estimate $a_j$ from the training data.
Next, define a map $\pi$ such that for $v  = (v_0, v_1, \dots , v_{k-1})^\top \in \RR^{k}$ we have $\pi(v) = (v_1, v_2, \dots, v_{k-1})^\top$.
For arbitrary $z \in \RR^{(n-1)}$ let
$$
B_z = \Big(z \mid \pi(a_1) \mid \pi(a_2) \mid \cdots \mid \pi(a_{k-1}) \Big) \in \RR^{(k-1) \times k}.
$$
Let $F_z = B_zA^{-1}$ and note that $F_z$ represents the linear transformation which maps the $j$-th row of
$A$ to the $j$-th row of $B_z$. We call $F_z$ a {\em filtration matrix}. Let
$$
W_z = F_z W \in \RR^{(k-1) \times p}.
$$
Finally, we define a new classifier
$$
h_z : x \mapsto W_z \cdot f(x).
$$
Our unlearning operation is thus
$$\mathfrak D_z:
\begin{cases}
	\mathcal H \to \mathcal H_{\lnot \mathcal C}\\
	h \mapsto h_z.
\end{cases}
$$
We call $\mathfrak D_z$ a {\em linear filtration}.

Note that $\mathfrak D_z$ replaces
$W$ by $W_z = F_z W$, hence after applying $\mathfrak D_z$ we may delete
$W$. This means that, even though our unlearning operation essentially
filters
the outputs	of the original classifier, the linearity of the filtering
operations
allows us to absorb the filter
into the classifier. This is an important feature in a situation were it may longer be
permissible to store the original classifier. 

So how do we choose $z \in \RR^{k-1}$?

{\bf $\blacktriangleright$ Naive unlearning. $~$}
$z = \pi(a_0)$. This gives $F_z = (0 \mid I_{k-1})$,
i.e. $F_z = \pi$. Intuitively, we may think of this choice as simply
cutting the output unit associated with $\mathcal C$ out off the
classifier. We call this unlearning operation the {\em naive method} and will
use it as a baseline to measure the improvements other methods provide.

\begin{figure}
	\vskip 0.1in
	\begin{center}
		\centerline{\includegraphics[width=\columnwidth]{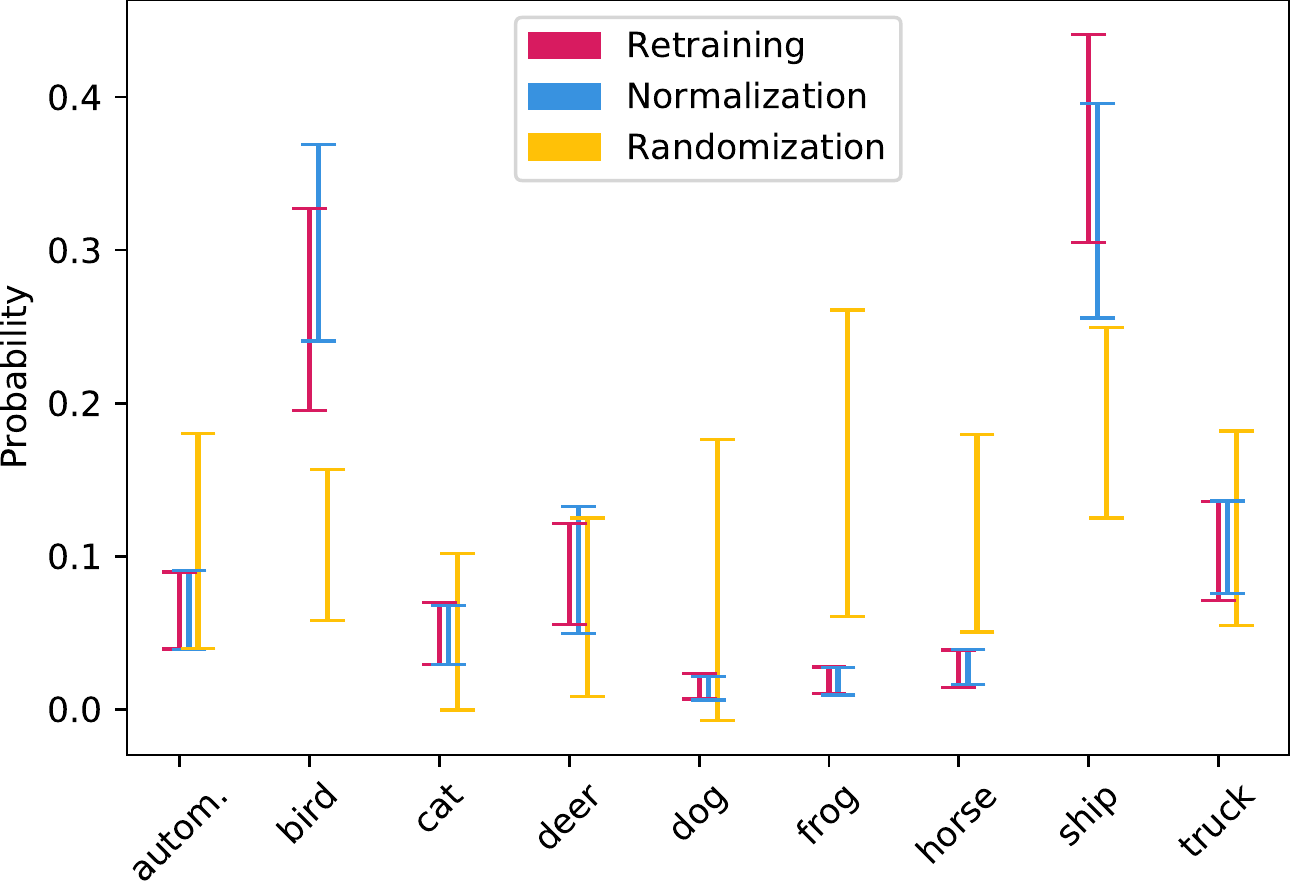}}
	\end{center}
	\vskip -0.3in
	\caption{
		The probability distribution predicted for class
		``airplane'' after its unlearning by either normalization or
		randomization from models trained on CIFAR-10,
		compared to models retrained without class ``airplane''.
		The bars are centered around the mean
		and have length of the standard deviation, over $100$ models.
	}
	\label{fig:misclassification}
	\vskip 0.2in
\end{figure}

{\bf $\blacktriangleright$ Normalization. $~$}
$$z = \pi(a_0) - \frac{1}{k-1} \sum_{1 \leq i < k} (a_0)_i
+ \frac{1}{(k-1)^2} \sum_{1 \leq i,j < k} (a_j)_i.$$
This means we shift $\pi(a_0)$ such that its mean becomes
the mean of the remaining rows of $B_z$.
The intuition behind this choice of $z$ is that we would like inputs of the
class in $\mathcal C$ to be misclassified in a ``natural'' way. We base
this method on the assumption that the values in $\pi(a_0)$ encode a natural distribution for predictions of the class in $\mathcal C$. However, we expect the values in $\pi(a_0)$ to be unnaturally low
(absolutely speaking),
hence we shift them. 
We refer to this method as {\em normalization} or {\em normalizing filtration}.
It is the main method we propose.
Figure \ref{fig:misclassification} agrees with the intuition described above: the bars for
normalization line up nicely
with the bars of the retrained models,
while the bars of randomization do not.

For comparison we define two more methods. 

{\bf $\blacktriangleright$ Randomization. $~$}
$z \sim \mathcal N(0, I_{k-1})$. We sample $z$ from
a multivariate normal distribution.
We refer to this method as {\em randomization}.

{\bf $\blacktriangleright$ Zeroing. $~$}
$z = 0$.
We refer to this method as {\em zeroing}.	

\subsection{Computational Complexity of $\mathfrak D_z$}
To find $\mathfrak D_z$ we need to compute the following:
\begin {enumerate*}[label=\textbf{\arabic*})]
\item
$A$, i.e. the expected predictions for all $k$ classes;
\item
$A^{-1}$, the inversion of a $(k \times k)$-Matrix;
\item
$z$, in case $z$ has a non-trivial definition (e.g.
computing a sample of $\mathcal N$);
\item
$F_z$, the multiplication of a $((k-1) \times k)$ with a
$(k \times k)$-Matrix;
\item
$W_z$, the multiplication of a $((k-1) \times k)$ with a
$(k \times p)$-Matrix.
\end {enumerate*} 

In practice {\bf 1)} will incur the majority of the computational costs,
while {\bf 2)}-{\bf 5)} will be negligible.
Thus, if we estimate $A$ by predicting $\ell$ samples per class
the computational complexity of finding $\mathfrak D_z(h)$ is
$\ell \cdot k $ times the complexity of computing a prediction of
$h$.	
We find that $\mathfrak D_z$ is robust in respect to the quality of
the estimation of $A$, hence a small amount of sample points per class
suffice, see section \ref{sec:results}.

Note that the costs incurred by linear filtration are virtually the same for several concurrent deletion requests: we just need to change $B_z$ appropriately, e.g. if we want to delete class $3$ in addition to class $0$ we need to replace $\pi(a_3)$ in column 3 with some $z_3$. The only additional costs incurred are the computation of $z_3$ (which for
our proposed methods are negligible).

\definecolor{seencolor}{RGB}{177,206,230}
\definecolor{notseencolor}{RGB}{255,241,200}
\definecolor{bincolor}{RGB}{243,145,145}
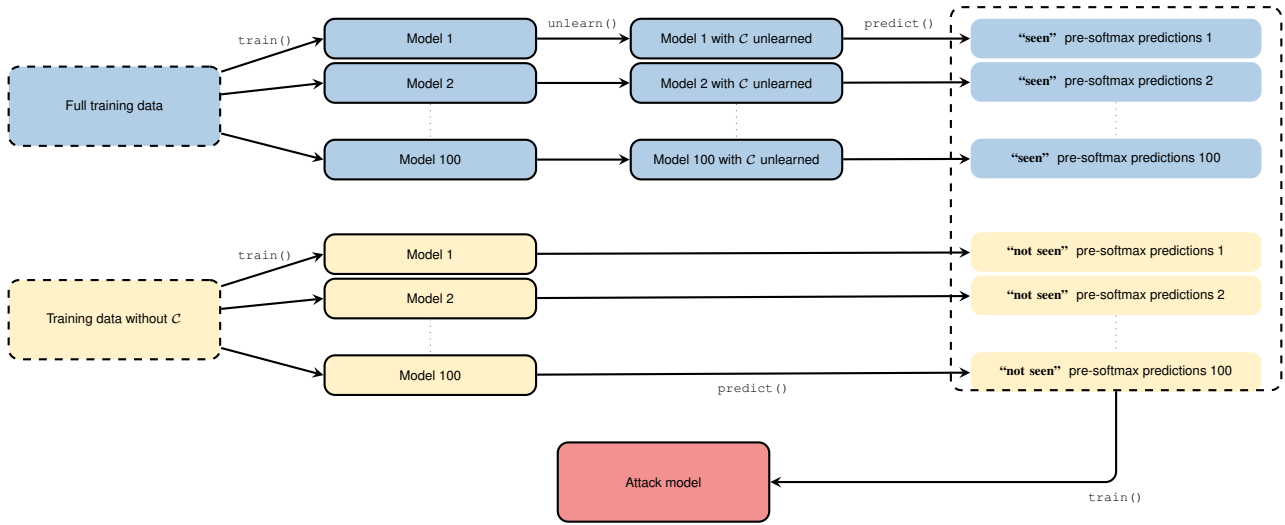
\begin{figure*}[ht]
	\vskip 0.1in
	\begin{center}
		\begin{tikzpicture}
		
		\draw (0,0) node[draw, thick, fill=seencolor, rounded corners,font=\sffamily\tiny,
		minimum height=1.5em, minimum width=8em]
		(m1) {Model 1};
		
		\draw node[draw, thick, below = .1em of m1, fill=seencolor, rounded corners,font=\sffamily\tiny,
		minimum height=1.5em, minimum width=8em]
		(m2) {Model 2};
		
		\draw node[draw, thick, below = 3em of m1, fill=seencolor, rounded corners,font=\sffamily\tiny,
		minimum height=1.5em, minimum width=8em]
		(m100) {Model 100};

		\draw node[draw, thick, right = 3.5em of m1, fill=seencolor, rounded corners,font=\sffamily\tiny,
		minimum height=1.5em, minimum width=8em]
		(n1) {Model 1 with $\mathcal C$ unlearned};
		
		\draw node[draw, thick, below = .1em of n1, fill=seencolor, rounded corners,font=\sffamily\tiny,
		minimum height=1.5em, minimum width=8em]
		(n2) {Model 2 with $\mathcal C$ unlearned};
		
		\draw node[draw, thick, below = 3em of n1, fill=seencolor, rounded corners,font=\sffamily\tiny,
		minimum height=1.5em, minimum width=8em]
		(n100) {Model 100 with $\mathcal C$ unlearned};

		\draw node[right = 4.8em of n1, fill=seencolor, rounded corners,font=\sffamily\tiny,
		minimum height=1.5em, minimum width=11em]
		(p1) {{\bf ``seen'' } pre-softmax predictions 1};
		
		\draw node[below = .1em of p1, fill=seencolor, rounded corners,font=\sffamily\tiny,
		minimum height=1.5em, minimum width=11em]
		(p2) {{\bf ``seen'' } pre-softmax predictions 2};
		
		\draw node[below = 3em of p1, fill=seencolor, rounded corners,font=\sffamily\tiny,
		minimum height=1.5em, minimum width=11em]
		(p100) {{\bf ``seen'' } pre-softmax predictions 100};

		\draw node[below = 2em of p100, fill=notseencolor, rounded corners,font=\sffamily\tiny,
		minimum height=1.5em, minimum width=11em]
		(q1) {{\bf ``not seen'' } pre-softmax predictions 1};
		
		\draw node[below = .1em of q1, fill=notseencolor, rounded corners,font=\sffamily\tiny,
		minimum height=1.5em, minimum width=11em]
		(q2) {{\bf ``not seen'' } pre-softmax predictions 2};
		
		\draw node[below = 3em of q1, fill=notseencolor, rounded corners,font=\sffamily\tiny,
		minimum height=1.5em, minimum width=11em]
		(q100) {{\bf ``not seen'' } pre-softmax predictions 100};

		\draw node[draw, thick, below = 2em of m100, fill=notseencolor, rounded corners,font=\sffamily\tiny,
		minimum height=1.5em, minimum width=8em]
		(d1) {Model 1};
		
		\draw node[draw, thick, below = .1em of d1, fill=notseencolor, rounded corners,font=\sffamily\tiny,
		minimum height=1.5em, minimum width=8em]
		(d2) {Model 2};
		
		\draw node[draw, thick, below = 3em of d1, fill=notseencolor,
		rounded corners,font=\sffamily\tiny,
		minimum height=1.5em, minimum width=8em]
		(d100) {Model 100};

		\draw (-4.2,-.9) node[draw, dashed, thick, fill=seencolor,
		rounded corners,font=\sffamily\tiny,
		minimum height=3em, minimum width=8em]
		(data1) {Full training data};
		
		\draw node[draw, dashed, thick, below = 5em of data1,
		fill=notseencolor, rounded corners,font=\sffamily\tiny,
		minimum height=3em, minimum width=8em]
		(data2) {Training data without $\mathcal C$};
		
		\draw node[below = -2em of p1, draw, dashed,thick,minimum height=14.5em,
		minimum width=12.5em, rounded corners]
		(pdata) {};
		
		\draw (3.1,-5.9)node[draw, thick,
		fill=bincolor, rounded corners,font=\sffamily\tiny,
		minimum height=3em, minimum width=8em]
		(bc) {Attack model};

		\draw[->, black!100, thick, >=stealth] (data1) -- (m1.west)
		node[midway,above,black]{\tiny\texttt{train()}$\quad$};
		\draw[->, black!100, thick, >=stealth] (data1) -- (m2.west);
		\draw[->, black!100, thick, >=stealth] (data1) -- (m100.west);
		
		\draw[->, black!100, thick, >=stealth] (data2) -- (d1.west)
		node[midway,above,black]{\tiny\texttt{train()}$\quad$};
		\draw[->, black!100, thick, >=stealth] (data2) -- (d2.west);
		\draw[->, black!100, thick, >=stealth] (data2) -- (d100.west);

		\draw[dotted, black!50] (m2) -- (m100);
		\draw[dotted, black!50] (d2) -- (d100);
		
		\draw[dotted, black!50] (p2) -- (p100);
		\draw[dotted, black!50] (q2) -- (q100);
		
		\draw[dotted, black!50] (n2) -- (n100);
		
		\draw[->, black!100, thick, >=stealth] (m1) -- (n1)
		node[midway,above,black]{\tiny\texttt{unlearn()}};
		\draw[->, black!100, thick, >=stealth] (m2) -- (n2);
		\draw[->, black!100, thick, >=stealth] (m100) -- (n100);

		\draw[->, black!100, thick, >=stealth] (n1) -- (p1)
		node[midway,above,black]{\tiny\texttt{predict()$\quad$}};
		\draw[->, black!100, thick, >=stealth] (n2) -- (p2);
		\draw[->, black!100, thick, >=stealth] (n100) -- (p100);
		
		\draw[->, black!100, thick, >=stealth] (d1) -- (q1);
		\draw[->, black!100, thick, >=stealth] (d2) -- (q2);
		\draw[->, black!100, thick, >=stealth] (d100) -- (q100)
		node[midway,below,black]{\tiny\texttt{predict()}};

		\draw[->, black!100, thick, >=stealth, rounded corners] (pdata.south) |- (bc.east)
		node[midway, below, black] {\tiny\texttt{train()}};
		\end{tikzpicture}
		\vskip 0.0in
		\caption{
			Experimental setup:
			On the full training data we
			train $100$ models by $\mathbf A\ =\ $\texttt{train()}.
			To these models we then apply an unlearning operation
			$\mathfrak D\ =\ $\texttt{unlearn()}. We then \texttt{predict()}
			our test data for each of these models and label these
			predictions {\bf ``seen''}.
			On the training data with $\mathcal C$ removed we train
			$100$ models by $\mathbf A_{\lnot \mathcal C}\ =\ $\texttt{train()}.
			We then \texttt{predict()}
			our test data for each of these models and label
			these predictions {\bf ``not seen''}.
			Finally, we use all labeled predictions as the training/test data of a binary classifier $b$,
			which we employ as our ``attack model''. We interpret low
			test accuracy of $b$ as evidence for good performance
			of a weak unlearning operation.
		}
		\label{fig:binary_classification}
	\end{center}
	\vskip 0.2in
\end{figure*}

\section{Experiments}
\label{sec:experiments}

\subsection{Evaluation method}
\label{sec:evaluation}
By definition \ref{dfn:weak_deletion} the quality of a weak unlearning operation
depends on the similarity of the resulting distributions $P(\mathbf L_{\mathrm{seen}})$ and $P(\mathbf L_{\lnot \mathrm{seen}})$.
We begin by defining a divergence measure for distributions based on the 
Bayes error rate. This then motivates us to empirically evaluate the performance of the unlearning operations proposed in
section~\ref{sec:method} by training binary classifiers on the pre-softmax outputs of our models.

{\bf Classifier advantage. $~$}
Let $\mathbf B$ be a Bernoulli random variable uniformly taking values in $\{0,1\}$. Let $$\mathbf U = \mathbf L_{\mathrm{seen}} \cdot (1 - \mathbf B) + \mathbf L_{\lnot \mathrm{seen}} \cdot \mathbf B$$
be the mixture of $\mathbf L_{\mathrm{seen}}$ and $\mathbf L_{\lnot \mathrm{seen}}$.
Let
$$
b : \dom(\mathbf U) \to \{0,1\}
$$
be a binary classifier operating on the mixture $\mathbf U$
and define
\begin{align}
	\label{form:advantage}
	\alpha_{b} = 2 \big( \EE[
	P(\mathbf B = b(\mathbf U) \mid \mathbf U) 
	] - \frac{1}{2} \big).
\end{align}
We call $\alpha_{b}$ the {\em classifier advantage} of $b$. Intuitively
speaking, $\alpha_b$ is a measure for how good $b$ is at telling the
mixture $\mathbf U$ apart.
Let
$$
b^*: u \mapsto \argmax_{i < 2} P(\mathbf B = i \mid \mathbf U = u)
$$
be the Bayes optimal classifier
for $P(\mathbf U, \mathbf B)$ 
Then $\alpha_{b^*} \in [0,1]$ and it is a measure for the difference between $P(\mathbf L_{\mathrm{seen}})$ and $P(\mathbf L_{\lnot \mathrm{seen}})$,
based on how much better the Bayes optimal classifier $b^*$ performs than random guessing. A value of $\alpha_{b^*}$ close to $0$ indicates that $P(\mathbf L_{\mathrm{seen}})$ and $P(\mathbf L_{\lnot \mathrm{seen}})$ are similar.

{\bf Experimental setup. $~$}
Assume that $b^*$ can be approximated sufficiently well by a classifier
$b$ derived via a state-of-the-art binary classification algorithm. 
Then $\alpha_b$ approximates $\alpha_{b^*}$, hence we consider a low value of $\alpha_b$ to be evidence for the similarity of the distributions of
$\mathbf L_{\mathrm{seen}}$ and $\mathbf L_{\lnot \mathrm{seen}}$. Note that even if we do not believe that $b$ approximates
$b^*$ well, we may still use $\alpha_b$ as a relative performance measure for
different unlearning operations.

Drawing i.i.d. samples from $P(\mathbf L_{\mathrm{seen}})$ and $P(\mathbf L_{\lnot \mathrm{seen}})$ is
computationally expensive as it requires us to run the algorithms
$\mathbf A(D)$ respectively $\mathbf A_{\lnot \mathcal C}(D_{\lnot \mathcal C})$
for every sample point. In our experiments we thus take a pragmatic approach. We train 100 models which get to see the full training data
and $100$ models which get to see the training data with $\mathcal C$
removed, i.e. models
that unlearned by
retraining from scratch. We then apply our unlearning operation to each of the models
that got to see the full training data. For every single model we then
calculate the predictions for the full test data by that model,
without applying the softmax activations of the output layer.
We label the predictions made by models that originally got to see
the full training data with {\bf ``seen''} and predictions
made by models which never got to see the training data for $\mathcal C$ as
{\bf ``not seen''}. 

Finally, we train a binary classifier $b$ that given a prediction
attempts to predict its label {\bf``seen''} or {\bf``not seen''}.
For this purpose we use the predictions of $70$ models of either
category as training data and the predictions of the remaining $30$ models of either category as test data. Figure \ref{fig:binary_classification}
shows a schematic representation of our setup. In practice, we train
a separate binary classifier on the predictions of each class.

Note the subtle difference of our setup to the shadow model setup of \cite{Shokri2017},
where the authors ask their binary classifier (there ``attack model''): 
``Do these outputs look like they come from a member of the training set?'',
and hope for an accurate answer, such that their membership inference attack succeeds.
We ask our binary classifier: ``Do these outputs look like they come from a model that has seen $\mathcal C$?'',
and hope for an inaccurate answer, as we hope that our unlearning operation prevents the attack.

\subsection{Data}
{\bf MNIST $~$}
The MNIST dataset \cite{MNIST}
contains $70{,}000$ $28{\times}28$ images of handwritten
digits in $10$ classes,
with $7{,}000$ images per class, 
split into $60{,}000$ training and $10{,}000$ test images.

{\bf CIFAR-10 $~$}
The CIFAR-10 dataset \cite{CIFAR}
contains $60{,}000$ $32{\times}32{\times}3$ images in $10$ classes,
with $6{,}000$ images per class, 
split into $50{,}000$ training and $10{,}000$ test images.

{\bf AT\&T Faces $~$} The AT\&T Laboratories Cambridge Database of Faces \cite{ATTFACES}
contains $400$ $92{\times}112$ images of $40$ subjects, with $10$
images per subject.

\subsection{Network Architectures}

{\bf MLP $~$}
For the MNIST dataset, we evaluate our unlearning method on a multilayer perceptron (MLP), with one hidden layer of 50 units.

{\bf CNN $~$}
For the CIFAR-$10$ dataset, we evaluate our unlearning method on convolutional neural networks (CNNs). Our networks consist of two convolutional
(with $16$ respectively $32$ $3{\times}3$ filters) and two max-pooling layers, followed by a fully connected layer with $p$ units with
rectified linear activations and a softmax output layer.
We experiment with $p \in \{64,256,1024 \}$.
Note that our unlearning operation works by manipulating $W \in \RR^{k \times p}$, thus we want to investigate how the size of $W$ affects
unlearning performance.

{\bf ResNet $~$}
For the AT\&T faces dataset we evaluate our unlearning method
on a residual neural network architecture \cite{resnet}.
We use a convolutional layer (with $8$ $5{\times}5$ filters),
followed by 5 downsampling residual blocks (with $2^{i+3}$
$3{\times}3$ filters in the $i$-th block), followed by global
max-pooling and a softmax output layer.

\subsection{Results}
\label{sec:results}
We experiment with the following classification algorithms:
nearest neighbors ({NN}), random forests ({RF}), and AdaBoost ({AB}).
For the models trained on MNIST and CIFAR-10 we unlearn the first class
(of $10$) in the dataset (the digit ``zero'' and the class ``airplane'', respectively).
For the models trained on AT\&T Faces we unlearn the first $4$ individuals
(of $40$) in the dataset.

\begin{table}[t]
	\caption{\normalfont Classifier advantage for 1 unlearned and
		9 remaining classes, for MLPs trained on MNIST.
	}
	\label{tab:MLP50}
	\vskip 0.0in
	\begin{center}
		\begin{small}
			\begin{sc}
				\begin{tabular}{lccc}
					\toprule
					Unlearned & NN & RF & AB \\
					\midrule
					Naive 	& 0.593	& 0.609	& 0.641\\
					Normalization 	& 0.327	& 0.362	& 0.438\\
					\bottomrule
					&&&\\
					\toprule
					Remaining & NN & RF & AB \\
					\midrule
					Naive 	& 0.048	& 0.090	& 0.098\\
					Normalization 	& 0.041	& 0.089	& 0.095\\
					\bottomrule
				\end{tabular}
			\end{sc}
		\end{small}
	\end{center}
	\vskip 0.2in
\end{table}

\begin{table}[t]
	\caption{\normalfont Classifier advantage for 1 unlearned and
		9 remaining classes, for CNNs trained on CIFAR-$10$, with $p = 256$.
	}
	\label{tab:CNN256}
	\vskip 0.0in
	\begin{center}
		\begin{small}
			\begin{sc}
				\begin{tabular}{lccc}
					\toprule
					Unlearned  & NN & RF & AB \\
					\midrule
					Naive 	& 0.609	& 0.457	& 0.590\\
					Normalization 	& 0.146	& 0.110	& 0.093\\
					Randomization 	& 0.634	& 0.579	& 0.582\\
					Zeroing 	& 0.642	& 0.566	& 0.575\\
					\bottomrule
					&&&\\
					\toprule
					Remaining & NN & RF & AB \\
					\midrule						
					Naive 	& 0.115	& 0.080	& 0.109\\
					Normalization 	& 0.148	& 0.097	& 0.118\\
					Randomization 	& 0.416	& 0.279	& 0.230\\
					Zeroing 	& 0.421	& 0.276	& 0.219\\
					\bottomrule
				\end{tabular}
			\end{sc}
		\end{small}
	\end{center}
	\vskip 0.2in
\end{table}

\begin{table}[t]
	\caption{\normalfont Classifier advantage for 4 unlearned and
		36 remaining classes, for residual networks trained on AT\&T faces.
	}
	\label{tab:RES}
	\vskip 0.0in
	\begin{center}
		\begin{small}
			\begin{sc}
				\begin{tabular}{lccc}
					\toprule
					Unlearned  & NN & RF & AB \\
					\midrule
					Naive 	& 0.467	& 0.573	& 0.574\\
					Normalization 	& 0.381	& 0.454	& 0.462\\
					\bottomrule
					&&&\\
					\toprule
					Remaining & NN & RF & AB \\
					\midrule
					Naive 	& 0.149	& 0.266	& 0.245\\
					Normalization 	& 0.152	& 0.263	& 0.246\\
					\bottomrule
				\end{tabular}
			\end{sc}
		\end{small}
	\end{center}
	\vskip 0.2in
\end{table}	

Table \ref{tab:MLP50} shows classifier advantages 
(remember formula (\ref{form:advantage})) for the MLPs trained on MNIST. We observe
a significant decrease of advantage when unlearning by normalization compared to the naive method
for the unlearned class, and similar advantage for the remaining classes.
Table \ref{tab:CNN256} shows classifier advantages for the CNNs trained on CIFAR-10 (with $p=256$).
We observe a vast decrease of advantage when unlearning by normalization compared to the naive method
for the unlearned class, while randomization and zeroing do not provide such benefits.
Table \ref{tab:RES} shows classifier advantages for the residual networks trained on AT\&T Faces. We observe
a slight decrease of advantage when unlearning by normalization compared to the naive method
for the unlearned classes, and similar advantage for the remaining classes.

\begin{table}[t]
	\caption{\normalfont Classifier advantage for CNNs trained on CIFAR-10, 
		for different sample size $s$ per class,
		with $p=256$.
	}
	\label{tab:sample_size}
	\vskip 0.0in
	\begin{center}
		\begin{small}
			\begin{sc}
				\begin{tabular}{lcccc}
					\toprule						
					Unlearned & $s$ & NN & RF & AB \\
					\midrule
					Naive 	& - & 0.609	& 0.457	& 0.590\\
					Normalization & 10 	& 0.193	& 0.114	& 0.122\\
					Normalization & 100	& 0.156	& 0.095	& 0.111\\
					Normalization & 1000& 0.146	& 0.110	& 0.093\\
					\bottomrule
					&&&&\\
					\toprule
					Remaining & $s$ & NN & RF & AB \\
					\midrule
					Naive 	& - & 0.115	& 0.080	& 0.109\\
					Normalization  & 10 		& 0.205	& 0.142	& 0.171\\
					Normalization  & 100  	& 0.169	& 0.108	& 0.136 \\
					Normalization  & 1000 	& 0.148	& 0.097	& 0.118\\
					\bottomrule
				\end{tabular}
			\end{sc}
			
		\end{small}
	\end{center}
	\vskip 0.2in
\end{table}
{\bf Effect of sample size. $~$}
We investigate the effect of sample size for estimating the matrix of
mean predictions $A$ (recall section \ref{sec:weak_del_op}).
Table \ref{tab:sample_size} shows the results for CNNs
trained on CIFAR-$10$ (with $p = 256$).
Unlearning by normalization, compared to the naive method, we observe a strong decrease of advantage for the unlearned class
for a sample size of $10$ per class ($1\	and for a sample size of $100$ per class ($10\	we observe performance comparable to the estimation based on the
full test data.

\begin{table}[t]
	\caption{\normalfont Classifier advantage, for CNNs trained on CIFAR-10, for different numbers $p$ of units in the fully
		connected layer.
	}
	\label{tab:hidden_units}
	\vskip 0.0in
	\begin{center}
		\begin{small}
			\begin{sc}
				\begin{tabular}{lcccc}
					\toprule						
					Unlearned & $p$  & NN & RF & AB \\
					\midrule
					Naive & 64	& 0.290	& 0.321	& 0.380\\
					Normalization & 64	& 0.086	& 0.106	& 0.119\\
					\midrule
					Naive & 256	& 0.609	& 0.457	& 0.590\\
					Normalization & 256	& 0.146	& 0.110	& 0.093\\
					\midrule
					Naive & 1024& 0.627	& 0.485	& 0.603\\
					Normalization & 1024& 0.174	& 0.138	& 0.138\\
					\bottomrule
					&&&&\\
					\toprule
					Remaining & $p$ & NN & RF & AB \\
					\midrule
					Naive & 64	& 0.040	& 0.043	& 0.044\\
					Normalization & 64	& 0.044	& 0.040	& 0.045\\
					\midrule
					Naive & 256 & 0.115	& 0.080	& 0.109\\
					Normalization & 256 & 0.148	& 0.097	& 0.118\\
					\midrule
					Naive  &1024& 0.129	& 0.090	& 0.115\\
					Normalization &1024	& 0.174	& 0.104	& 0.129\\
					\bottomrule
				\end{tabular}
			\end{sc}
			
		\end{small}
	\end{center}
	\vskip 0.2in
\end{table}
{\bf Effect of $p$. $~$}
For the CNNs trained on CIFAR-$10$ we investigate
the effect of the number $p$ of units in the fully connected layer.
Table \ref{tab:hidden_units} shows that unlearning by normalization significantly decreases
classifier advantage for all tested values of $p$ on the unlearned class.
We observe a slight increase of advantage for the remaining classes,
that appears to get slightly worse with $p$ increasing. 	

\begin{table}[t]
	\caption{\normalfont The amount of labels changed in percent,
		when compared to the naive method, for 100 models trained on CIFAR-10,
		and $p = 256$.
		The ``all''-column reports the value for predictions of
		the entire test set,
		the ``unl.''-column for predictions of images of the unlearned class,
		and the ``cor.''-column for predictions of images of the remaining
		classes that are correctly predicted by the naive method.
		The ``acc.''-column shows the mean classification accuracy.
	}
	\label{tab:label_flipping}
	\vskip 0.0in
	\begin{center}
		\begin{small}
			\begin{sc}
				\begin{tabular}{lccccc}
					\toprule
					Deletion method & All & Unl. & Cor. & Acc \\
					\midrule						
					Naive & - & - & - & 69.2 \\
					Normalization & 0.0 & 0.0 & 0.0 & 69.2\\
					Randomization & 20.1 & 48.8 & 11.8 & 64.8 \\
					Zeroing & 18.3 & 45.4 & 10.3 & 65.7 \\
					\bottomrule
				\end{tabular}
			\end{sc}
		\end{small}
	\end{center}
	\vskip 0.2in
\end{table}

{\bf Effect on classification accuracy. $~$}
While implicit in definition \ref{dfn:weak_deletion}, an important consideration for the design of an unlearning operation is that
we do not want to decrease the performance of a classifier on the remaining classes.
Table \ref{tab:label_flipping} reports the test images
for which the most likely label predicted was changed by one of our unlearning methods, compared to
the naive method. Normalization did not change
any labels (and thus in particular non of the correct ones).

{\bf Summary of experimental results. $~$}
Normalization
\begin{enumerate*}[label=\textbf{\arabic*})]
	\item
	decreased
	classifier advantage on the unlearned classes in all three experiments
	\item
	showed robustness with regards to sample size for parameter estimation
	\item
	performed well for different dimensions of the feature space, and
	\item did not
	negatively affect classification accuracy.
\end{enumerate*}

\subsection{Model inversion}
\label{sec:model_inversion}
\begin{figure}
	\vskip 0.1in
	\begin{center}
		\centerline{\includegraphics[height=80mm]{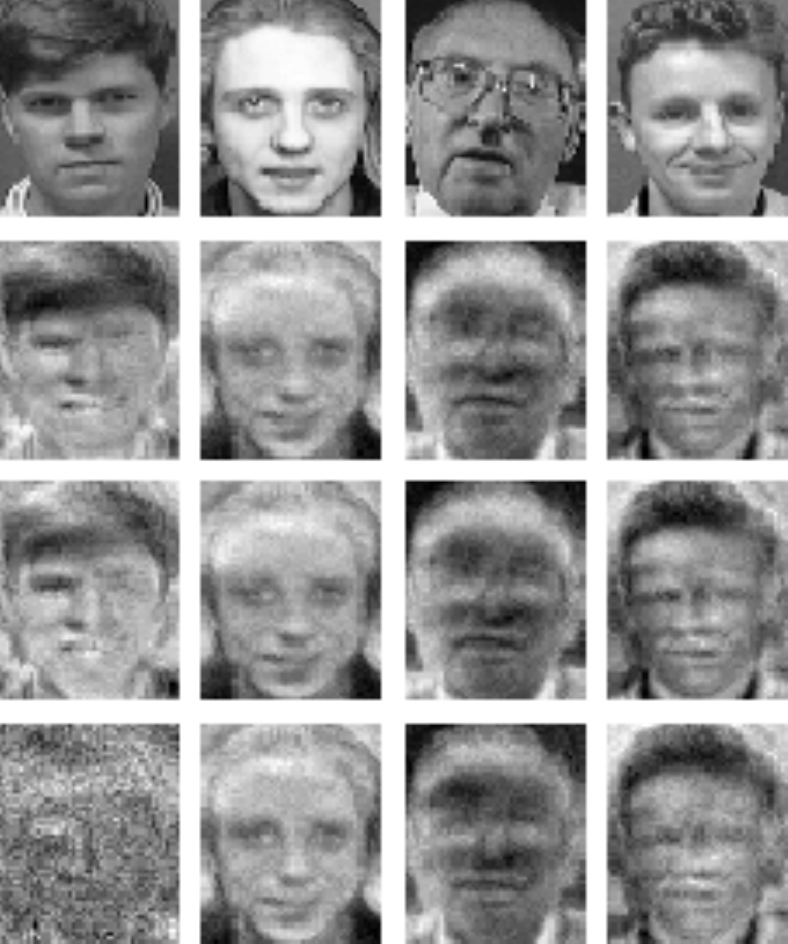}}
	\end{center}
	\vskip -0.2in
	\caption{
		Model inversion for a toy model trained on
		the AT\&T Faces dataset with 4 classes.
		The top row shows one training image of each class,
		the second row reconstructions of classes by model inversion,
		the third row reconstructions after naive unlearning
		of the class in the first column, the bottom row reconstructions
		after unlearning the class in the first column by
		{\em normalizing filtration}.
        See figure \ref{fig:att_inversion_full} for the remaining classes.
	}
	\label{fig:att_inversion2}
	\vskip 0.2in
\end{figure}

Figure \ref{fig:att_inversion1}
and figure \ref{fig:att_inversion2} show the results of a ``model inversion
attack'', i.e. gradient ascend on the input space for a model
trained on AT\&T Database of Faces.	
We use a neural network consisting
of two fully connected layers with $1000$ and $300$ units respectively with
sigmoid activations
and a softmax output layer.

Visually, naive unlearning barely affects
the quality of the reconstruction for any of the classes
(in particular not the unlearned class). On the other hand
the normalization method greatly disturbs the reconstruction of
the unlearned class, while barely affecting the remaining classes. 
In accordance with our discussion of model inversion in section
\ref{sec:related_work},
we thus interpret figure~\ref{fig:att_inversion1} as visual evidence
suggesting	a desirable effect of our normalization method on the
correlation	between input and output space represented by our model.
We leave a more detailed investigation of this phenomenon for future work.

\subsection{Random direction Kolmogorov–Smirnov statistics}
\label{sec:ks}
Given $\Phi \subseteq \RR^{k - |\mathcal C|}$, and having drawn samples $S_1$, $S_2$ from $P(\mathbf L_{\mathrm{seen}})$ and $P(\mathbf L_{\lnot \mathrm{seen}})$ respectively in the fashion described  in section
\ref{sec:evaluation}, we compute the statistic
\begin{align}
\label{form:ks}
\KS_\Phi(S_1, S_2) = \frac{1}{|\Phi|} \sum_{\phi \in \Phi}
\KS_{\textrm{two sample}}(\phi \cdot S_1, \phi \cdot S_2).
\end{align}

Here $KS_{\textrm{two sample}}$ denotes the two-sample Kolmogorov–Smirnov statistic and $\phi \cdot S$ is shorthand for
$\{\phi \cdot s : s \in S \} \subseteq \RR$.

Table \ref{tab:ks} reports the $\KS_\phi$ statistic for
$\Phi$ consisting of
$1000$ unit vectors pointing in uniform random directions.
We compare
naive unlearning to normalizing filtration for the
CIFAR-10 experiment employing CNNs ($p = 256$), and
the MNIST experiment employing MLPs.
As noted in section \ref{sec:evaluation} the samples we draw are not
i.i.d., hence we also report a baseline $\KS_\phi$ statistic computed
from two independent batches of 100 models trained without 
the class we unlearn.

The results paint a picture similar to section \ref{sec:results}.
In both experiments normalizing filtration decreases $\KS_\Phi$ for the
unlearned class, with somewhat better performance in the CIFAR-10 experiment, while the $\KS_\Phi$ remains unchanged for the
remaining classes.

\begin{table}[t]
	\caption{\normalfont $\KS_\Phi$ for $\Phi$ consisting of
		$1000$ unit vectors pointing in uniform random directions.
		(Lower values are better.)
	}
	\label{tab:ks}
	\vskip 0.0in
	\begin{center}
		\begin{small}
			\begin{sc}
				\begin{tabular}{lcc}
					\toprule						
					Unlearned & CIFAR-10 & MNIST \\
					\midrule
					Naive 	& .115 & .184 \\
					Normalization & .038 	& .1\\
					Baseline & .015	& .017	\\
					\bottomrule
					&&\\
					\toprule				
					Remaining & CIFAR-10 & MNIST \\
					\midrule
					Naive 	& .037 & .034 \\
					Normalization & .038 	& .034\\
					Baseline & .013	& .023	\\
					\bottomrule
				\end{tabular}
			\end{sc}
			
		\end{small}
	\end{center}
	\vskip 0.2in
\end{table}

\section{Discussion}
\label{sec:on_the_def}

In section \ref{sec:problem_def} we made a didactic choice to introduce the special hypothesis class $\mathcal H$ (which permitted absorption of the filtration operation)
before defining unlearning. It should however be clear how definitions \ref{dfn:deletion} and \ref{dfn:weak_deletion} are applicable
to any class of classifiers, and in the somewhat more common setting of
deletion requests of single data points.
We would like to emphasize our belief that in the light of the non-deterministic nature of many learning algorithms a probabilistic
definition of unlearning, such as chosen in \cite{Ginart2019} (there ``deletion'')
and our work, is necessary.

For contrast let us consider the definition of
unlearning in \cite{bourtoule2019} that asks
to find a model that ``\underline{could have} been obtained'' without looking
at the deleted training data. Since all models on discrete digital systems
necessarily have a finite parameter space we very much \underline{could} obtain
any model without looking at any data by guessing its parameters.
What happened here? 

Guessing a model's parameters can be considered
drawing a sample from a uniform distribution over the parameter space.
On the other hand a probabilistic definition such as \ref{dfn:deletion}
requires the distribution over the parameter space to be the same
as if we had run the original learning algorithm without using the deleted
data, which for reasonable learning algorithms is certainly not uniform.
If we would like to stick to an informal definition we should therefore say
that a model ``\underline{could have} been obtained, with \underline{reasonable likelihood}''.

We further conclude that it is an important consideration whether a definition
of unlearning makes sense when not read in a benevolent way (e.g. by a party whose interest in unlearning stems from of legal obligation).
In fact our definition of weak unlearning \ref{dfn:weak_deletion},
i.e. unlearning in a black-box sense, suffers from a similar issue.
A malicious way to define
a weak unlearning operations $\mathfrak D$ is the following: for any classifier
$h$, simply train a new classifier $h'$ without using the deleted data, then 
define $\mathfrak D(h) = h'' = (h,h')$, where $h''(x) = h'(x)$. 
The outputs of $\mathfrak D(h)$ look exactly like those of 
$h'$, thus $\mathfrak D$ is indeed an unlearning operation in the
black-box sense, however we never actually deleted $h$.
Let us thus emphasize that the weak definition of unlearning is only applicable when acting
in good faith.

\section{Conclusion}

\begin{figure*}[ht]
	\vskip 0.2in
	\centering
	\includegraphics[width=80mm]{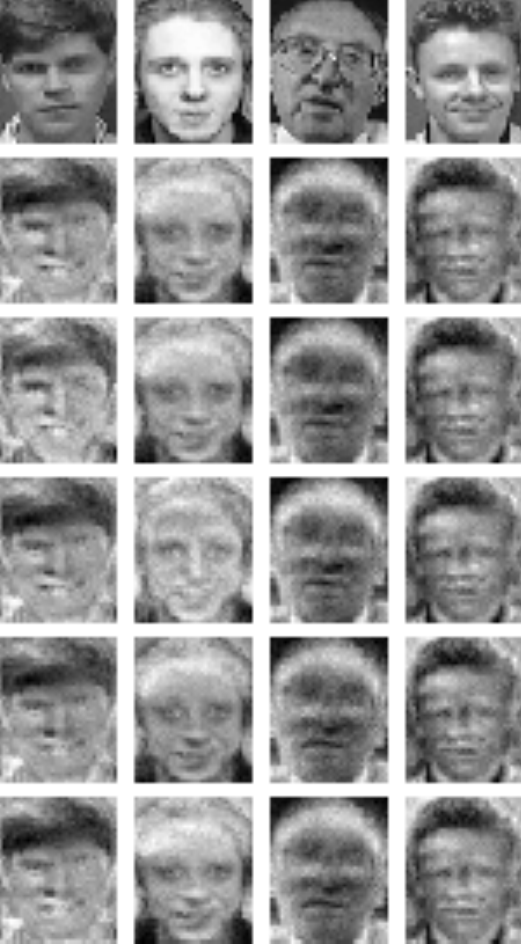}
	\hspace{.2in}
	\includegraphics[width=80mm]{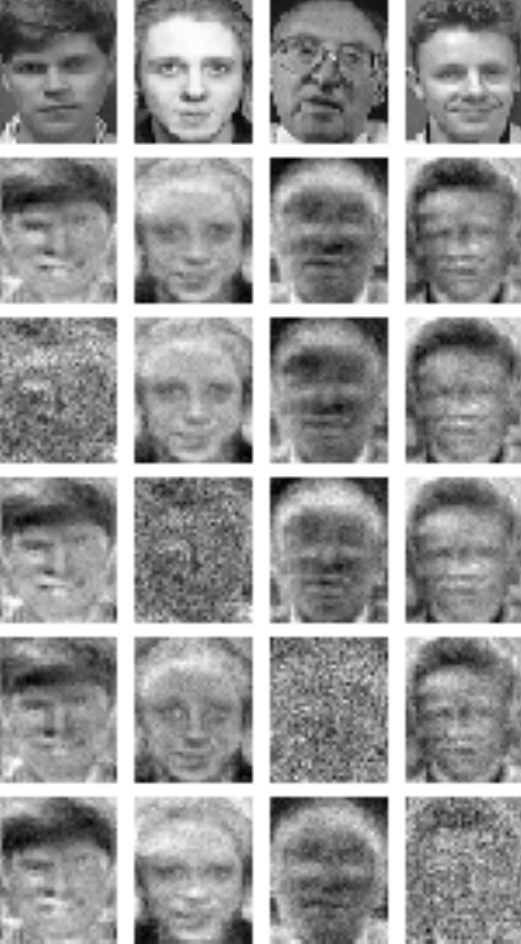}
	
	\vskip 0.2in
	\caption{
		Model inversion for a toy model trained on
		the AT\&T Faces dataset with 4 classes.
		On the left side we show results
		for naive unlearning, on the right side for
		{\em normalizing filtration}.
		On either side:
		The top row shows one training image of each class,
		the second row reconstructions of classes by model inversion,
		the $(i + 2)$-th row shows the reconstructions
		after unlearning the class in the $i$-th column
		by the respective unlearning operation.
	}
	
	\label{fig:att_inversion_full}
	\vskip -0.2in
\end{figure*}

We considered the problem of unlearning in a class-wide setting, for classifiers
predicting logits. We developed {\em normalizing filtration} as an unlearning
method, with compelling visual results (figure \ref{fig:att_inversion_full}).
These are backed up by good experimental results with regards
to to our proposed definition of {\em weak unlearning} and
our metrics of {\em classifier advantage} (\ref{form:advantage}) and
the $\KS_\Phi$ statistic (\ref{form:ks}).
We emphasize once again the black-box nature of weak unlearning,
entailing the black-box nature of our proposed unlearning operation.
While linear filtration can be absorbed
into the final layer of a classifier (for the hypothesis class considered), our approach remains somewhat limited with regards to its shallowness. 
In future work we hope to find methods that allow for deeper absorption,
thus hopefully leading to stronger privacy guarantees.
On the other hand our method's intuitive simplicity nicely complements
concurrent approaches such as \cite{guo2019, golatkar2019, golatkar2020}.
Another promising direction may be to enhance our method by some form of
shrinkage of the logits, in case the unlearned class constitutes
a large part of the misclassifications of one of the remaining classes.
Finally it should be noted that, while the ``right to be forgotten''
inspired our research, whether our approach is adequate in
this context is for legal scholars to decide.

\bibliographystyle{IEEEtran}
\bibliography{IEEEabrv,citations}

\clearpage 

\onecolumn
\appendix

More complete versions of tables from section
\ref{sec:results}.

\subsection{MNIST -- MLP experiment}

	\begin{table}[ht]
		\caption{\normalfont
	    Accuracy (in percent) and cross-entropy loss,
	    mean $\pm$ standard deviation, for 100 MLPs trained
	    on MNIST.
	    $p$ is the number of units in the fully connected
	    layer, i.e. the dimension of the feature space.
	    $s$ is the sample size per class, used to estimate
	    the mean predictions.
		}
		\label{tab:app-MNIST-acc}
		\vskip 0.15in
		\begin{center}
			\begin{small}
				\begin{sc}
					\begin{tabular}{lccc}
						\toprule
						Unlearned & NN & RF & AB \\
						\midrule
						Naive 	& 0.593	& 0.609	& 0.641\\
						Normalization 	& 0.327	& 0.362	& 0.438\\
						Randomization 	& 0.910	& 0.869	& 0.902\\
						Zeroing 	& 0.919	& 0.886	& 0.909\\
						\bottomrule
						&&&\\
						\toprule
						Remaining & NN & RF & AB \\
						\midrule
						Naive 	& 0.048	& 0.090	& 0.098\\
						Normalization 	& 0.041	& 0.089	& 0.095\\
						Randomization 	& 0.116	& 0.132	& 0.133\\
						Zeroing 	& 0.119	& 0.129	& 0.132\\
						\bottomrule
					\end{tabular}
				\end{sc}
			\end{small}
		\end{center}
		\vskip -0.1in
	\end{table}

\begin{table}[h]
	\caption{\normalfont
	    Classifier advantage for 1 unlearned and 9 remaining classes, for MLPs trained on MNIST.
	    $p$ is the number of units in the fully connected
	    layer, i.e. the dimension of the feature space.
	    $s$ is the sample size per class, used to estimate
	    the mean predictions.}
	\label{tab:app-MNIST-adv}
	\vskip 0.15in
	\begin{center}
		\begin{small}
			\begin{sc}
				\begin{tabular}{lcc}
					\toprule
					& Accuracy & Loss\\
					\midrule
					Before unlearning & 97.1 $\pm$ 0.19 & 0.10 $\pm$ 0.01 \\
					Baseline model & 97.1 $\pm$ 0.18 & 0.09 $\pm$ 0.01 \\
					Naive & 97.1 $\pm$ 0.20 & 0.10 $\pm$ 0.01 \\
					Normalization & 97.1 $\pm$ 0.20 & 0.10 $\pm$ 0.01 \\
					Randomization & 96.7 $\pm$ 0.33 & 0.11 $\pm$ 0.01 \\
					Zeroing & 96.7 $\pm$ 0.30 & 0.11 $\pm$ 0.01 \\
					\bottomrule
				\end{tabular}
			\end{sc}
		\end{small}
	\end{center}
	\vskip -0.1in
\end{table}	

\clearpage
\subsection{CIFAR10 -- CNN experiment}
	
	\begin{table}[h]
		\caption{\normalfont
	    Classifier advantage for 1 unlearned and 9 remaining classes, for CNNs trained on CIFAR-10.
	    $p$ is the number of units in the fully connected
	    layer, i.e. the dimension of the feature space.
	    $s$ is the sample size per class, used to estimate
	    the mean predictions.
		}
		\label{tab:app-CIFAR10-adv}
		\vskip 0.15in
		\begin{center}
			\begin{small}
				\begin{sc}
					\begin{tabular}{lccccc}
						\toprule
						Unlearned & $p$ & $s$ & NN & RF & AB \\
						\midrule
                        Naive  & 64 & - 	& 0.290	& 0.331	& 0.380\\
                        Normalization & 64 & 10 	& 0.113	& 0.135	& 0.137\\
                        Normalization & 64 & 100 	& 0.089	& 0.105	& 0.129\\
                        Normalization  & 64 & - 	& 0.086	& 0.106	& 0.119\\
                        Randomization  & 64 & - 	& 0.547	& 0.487	& 0.528\\
                        Zeroing  & 64 & - 	& 0.599	& 0.534	& 0.560\\
                        \midrule
                        Naive  & 256 & - 	& 0.609	& 0.457	& 0.590\\
                        Normalization  & 256 & 10 	& 0.193	& 0.114	& 0.122\\
                        Normalization  & 256 & 100 	& 0.156	& 0.095	& 0.111\\
                        Normalization  & 256 & - 	& 0.146	& 0.110	& 0.093\\
                        Randomization  & 256 & - 	& 0.634	& 0.579	& 0.582\\
                        Zeroing  & 256 & - 	& 0.642	& 0.566	& 0.575\\
                        \midrule
                        Naive  & 1024 & - 	& 0.627	& 0.485	& 0.603\\
                        Normalization & 1024 & 10 	& 0.230	& 0.164	& 0.179\\
                        Normalization  & 1024 & 100	& 0.216	& 0.157	& 0.172\\
                        Normalization  & 1024 & - 	& 0.174	& 0.138	& 0.138\\
                        Randomization  & 1024 & - 	& 0.741	& 0.637	& 0.682\\
                        Zeroing  & 1024 & - 	& 0.746	& 0.642	& 0.680\\
						\bottomrule
						&&&\\
						\toprule
						Remaining & $p$ & $s$ &  NN & RF & AB \\
						\midrule
                        Naive   & 64 & - 	& 0.040	& 0.043	& 0.044\\
                        Normalization    & 64 & 10 	& 0.072	& 0.075	& 0.080\\
                        Normalization   & 64 & 100 	& 0.047	& 0.044	& 0.047\\
                        Normalization   & 64 & - 	& 0.044	& 0.040	& 0.045\\
                        Randomization   & 64 & - 	& 0.240	& 0.203	& 0.174\\
                        Zeroing   & 64 & - 	& 0.246	& 0.204	& 0.163\\
                        \midrule
                        Naive   & 256 & - 	& 0.115	& 0.080	& 0.109\\
                        Normalization   & 256 & 10 	& 0.205	& 0.142	& 0.171\\
                        Normalization  & 256 & 100 	& 0.169	& 0.108	& 0.136\\
                        Normalization   & 256 & - 	& 0.148	& 0.097	& 0.118\\
                        Randomization   & 256 & - 	& 0.416	& 0.279	& 0.230\\
                        Zeroing   & 256 & - 	& 0.421	& 0.276	& 0.219\\
                        \midrule
                        Naive   & 1024 & - 	& 0.129	& 0.090	& 0.115\\
                        Normalization  & 1024 & 10 	& 0.242	& 0.149	& 0.182\\
                        Normalization & 1024 & 100 	& 0.217	& 0.128	& 0.146\\
                        Normalization   & 1024 & - 	& 0.174	& 0.097	& 0.118\\
                        Randomization   & 1024 & - 	& 0.506	& 0.309	& 0.249\\
                        Zeroing   & 1024 & - 	& 0.508	& 0.310	& 0.251\\
						\bottomrule
					\end{tabular}
				\end{sc}
			\end{small}
		\end{center}
		\vskip -0.1in
	\end{table}

\begin{table}[h]
	\caption{\normalfont
	    Accuracy (in percent) and cross-entropy loss,
	    mean $\pm$ standard deviation, for 100 CNNs
	    trained on CIFAR10.
	    $p$ is the number of units in the fully connected
	    layer, i.e. the dimension of the feature space.
	    $s$ is the sample size per class, used to estimate
	    the mean predictions.
	}
	\label{tab:app-CIFAR10-acc}
	\vskip 0.15in
	\begin{center}
		\begin{small}
			\begin{sc}
				\begin{tabular}{lcccc}
					\toprule
					& $p$ & $s$ & Accuracy & Loss\\
					\midrule
					Before unlearning & 64 & - & 65.1 $\pm$ 1.01 & 1.01 $\pm$ 0.03 \\
                    Retraining & 64 & - & 66.0 $\pm$ 1.06 & 0.96 $\pm$ 0.03 \\
                    Naive & 64 & - & 66.4 $\pm$ 1.04 & 0.95 $\pm$ 0.03 \\
                    Normalization & 64 & 10 & 66.4 $\pm$ 1.04 & 0.95 $\pm$ 0.03 \\
                    Normalization & 64 & 100 & 66.4 $\pm$ 1.04 & 0.95 $\pm$ 0.03 \\
                    Normalization & 64 & - & 66.4 $\pm$ 1.04 & 0.95 $\pm$ 0.03 \\
                    Randomization & 64 & - & 61.1 $\pm$ 2.08 & 1.17 $\pm$ 0.10 \\
                    Zeroing & 64 & - & 62.3 $\pm$ 1.76 & 1.12 $\pm$ 0.09 \\
                    \midrule
                    Before unlearning & 256 & - & 68.1 $\pm$ 0.85 & 0.93 $\pm$ 0.02 \\
                    Retraining & 256 & - & 69.2 $\pm$ 0.80 & 0.89 $\pm$ 0.02 \\
                    Naive & 256 & - & 69.2 $\pm$ 0.83 & 0.88 $\pm$ 0.02 \\
                    Normalization & 256 & 10 & 69.2 $\pm$ 0.83 & 0.88 $\pm$ 0.02 \\
                    Normalization & 256 & 100 & 69.2 $\pm$ 0.83 & 0.88 $\pm$ 0.02 \\
                    Normalization & 256 & - & 69.2 $\pm$ 0.83 & 0.88 $\pm$ 0.02 \\
                    Randomization & 256 & - & 64.9 $\pm$ 1.76 & 1.07 $\pm$ 0.08 \\
                    Zeroing & 256 & - & 65.7 $\pm$ 1.40 & 1.03 $\pm$ 0.06 \\
                    \midrule
                    Before unlearning & 1024 & - & 69.7 $\pm$ 0.81 & 0.93 $\pm$ 0.03 \\
                    Retraining & 1024 & - & 70.5 $\pm$ 0.82 & 0.88 $\pm$ 0.03 \\
                    Naive & 1024 & - & 70.7 $\pm$ 0.82 & 0.88 $\pm$ 0.03 \\
                    Normalization & 1024 & 10 & 70.7 $\pm$ 0.82 & 0.88 $\pm$ 0.03 \\
                    Normalization & 1024 & 100 & 70.7 $\pm$ 0.82 & 0.88 $\pm$ 0.03 \\
                    Normalization & 1024 & - & 70.7 $\pm$ 0.82 & 0.88 $\pm$ 0.03 \\
                    Randomization & 1024 & - & 67.1 $\pm$ 1.55 & 1.06 $\pm$ 0.08 \\
                    Zeroing & 1024 & - & 67.6 $\pm$ 1.16 & 1.03 $\pm$ 0.05 \\
					\bottomrule
				\end{tabular}
			\end{sc}
		\end{small}
	\end{center}
	\vskip -0.1in
\end{table}	

\end{document}